\documentclass{article}

\usepackage[preprint]{corl_2026} 

\title{PHASOR: Phase-Anchored Universal Action Representations for Humanoid Embodiments}

\author{
\textbf{Kihyun Kim}$^{1,2\ast}$ \quad
\textbf{Chaeyun Kim}$^{1,2\ast\dagger}$ \quad
\textbf{Jongho Shin}$^{3}$ \quad
\textbf{Taeyoun Kwon}$^{2,4}$ \\[0.3em]
\textbf{Junghyun Kim}$^{2,5}$ \quad
\textbf{Mijin Koo}$^{1,2}$ \quad
\textbf{Haon Park}$^{1,2\dagger}$ \\[0.6em]
$^{1}$AIM Intelligence \quad
$^{2}$Seoul National University \quad
$^{3}$LG Electronics \\[0.2em]
$^{4}$MAUM AI \quad
$^{5}$OpenMind
\\[-2em]
}

\newcommand{\blfootnote}[1]{%
  \begingroup
  \renewcommand\thefootnote{}\footnote{#1}%
  \addtocounter{footnote}{-1}%
  \endgroup
}

\usepackage{graphicx}
\usepackage{booktabs}
\usepackage{amssymb}
\usepackage{amsfonts}
\usepackage{mathtools}
\usepackage{caption}
\usepackage{subcaption}
\usepackage{wrapfig}
\usepackage[most]{tcolorbox} 
\usepackage{multirow}
\usepackage{pifont}
\usepackage{enumitem}
\usepackage[dvipsnames,table]{xcolor}
\usepackage{pifont}
\usepackage{url} 

\usepackage[capitalise]{cleveref}
\crefname{table}{Tab.}{Tabs.}
\Crefname{table}{Tab.}{Tabs.}
\crefname{section}{Sec.}{Secs.}
\Crefname{section}{Sec.}{Secs.}
\Crefname{appendix}{App.}{Apps.}

\begin{document}
\maketitle
\vspace{-1.8em}
\blfootnote{$^{\ast}$Equal contribution. \quad $^{\dagger}$Corresponding author.}

\begin{abstract}
    Learning a good action embedding space is fundamental to scalable robot policy learning, yet existing methods treat action latents as task-specific intermediates rather than first-class representations. The resulting latents are unstructured, embodiment-specific, and weakly tied to motion semantics, limiting interpretability, controllability, and transferability across robots. We position the action embedding space itself as a first-class design target, with downstream policy quality emerging from representation quality. Exploiting motion's intrinsic periodicity, we factorize it into a phase manifold that captures cyclic structure via FFT-parametric coefficients, together with a pose branch that conditions the manifold on non-periodic configuration detail. Combined with motion-semantic distillation, this factorized structure yields a cross-embodiment motion manifold that is interpretable and embodiment-agnostic by design. Anchoring multiple humanoid robots to a shared human-pretrained manifold then produces a unified action embedding space across diverse platforms, achieving strong cross-embodiment retrieval and consistent gains on downstream robot tasks.
\end{abstract}

\section{Introduction}
\label{sec:intro}

The rapid diversification of humanoid robot platforms has created a pressing need for action representations that generalize across embodiments. Modern robot policies, from vision-language-action models~\citep{zitkovich2023rt,black2024pi_0,bjorck2025gr00t} to diffusion-based controllers~\citep{chi2025diffusion,luo2025sonic}, all build on learned action spaces as the intermediate signal between perception and motor execution. In cross-embodiment settings, this space becomes the primary bridge through which knowledge transfers from one body to another, making its quality a bottleneck for generalization.

Yet current action encoders~\citep{ye2025latent,mete2024quest,team2024octo,bu2025univla,kim2025openvla} treat motion as a generic sequence of per-frame joint states, with no motion-specific structure. This leads to two failures. 
First, the representation entangles with the morphology of the training robot, so the same behavior on a different humanoid maps to a disjoint region and cannot transfer. Second, the latent space lacks temporal organization, and proximity within it does not reflect proximity in time. Thus its dimensions carry no interpretable meaning, leaving trajectories hard to analyze or control.

To overcome these limitations, we exploit the intrinsic physical structure of motion. Unlike text or images, motion is fundamentally a superposition of periodic components. During walking, the legs oscillate at roughly 1\,Hz, the arms swing in antiphase, and the trunk rotates more slowly. Decomposing motion into amplitude, frequency, and phase shift~\citep{starke2022deepphase} separates two quantities that unstructured encoders conflate. Phase encodes \emph{where in the action} the body is, while amplitude and frequency encode \emph{what action} is performed. We hypothesize that this \textbf{periodic structure is intrinsic} to the behavior rather than to the body that produces it, and can therefore \textbf{serve as a shared substrate for aligning} action embeddings across robots~\citep{li2024walkthedog}.

Realizing this phase manifold as a cross-embodiment action space requires solving two problems. First, existing approaches~\citep{starke2022deepphase,li2024walkthedog,ji2025pomp,pegoraro2025funphase} compress the whole body into a single representation, blurring the distinct periodicities of individual joints. We instead decompose the phase encoding by body part, preserving each limb's periodic signature as a natural unit of correspondence across kinematics. 
Second, encoding motion through velocity alone captures how the body moves but not where it is; so motions with similar oscillation but different poses (e.g., upright vs.\ crouched walking) collapse to nearby points. We resolve this with a separate pose stream that supplies non-periodic context to the decoder, leaving $\mathbf{P}(\tau)$ uncontaminated.


With this clean manifold established, the remaining challenge is extending it across embodiments. Since periodic structure belongs to the behavior rather than the skeleton, and human data far outnumbers fragmented robot data, we anchor the shared manifold in human motion. Here, each new robot joins through a lightweight adapter rather than relearning the manifold from scratch.


We propose \textbf{PHASOR} (\textbf{PH}ase \textbf{A}ction \textbf{S}pace for cross-emb\textbf{O}diment \textbf{R}epresentation), a framework built on the periodic structure of motion. A per-body-part phase encoder extracts periodic parameters from joint velocities to assemble a phase manifold $\mathbf{P}(\tau)$, while a complementary pose branch supplies non-periodic context through FiLM~\citep{perez2018film} without leaking into $\mathbf{P}(\tau)$. The human-pretrained encoder acts as a shared anchor, and each robot joins through a lightweight adapter. An alignment head then projects per-embodiment manifolds onto a unit sphere and pulls semantically equivalent motions together, guided by a pretrained semantic motion prior~\citep{li2025lamp}.

We validate this design on one human and four humanoid robots along two axes, \textbf{\emph{embedding quality}} and \textbf{\emph{downstream utility}}. The phase manifold yields a tightly aligned cross-embodiment space, reaching over 88\% cross-embodiment retrieval accuracy and surpassing unstructured and quantized baselines. 
As a learned representation, it further improves motion imitation, teleoperation, and reinforcement learning, confirming that motion-intrinsic periodicity is an effective design prior.

Our contributions are:
\vspace{-2pt}
\begin{enumerate}[itemsep=0pt, topsep=2pt, parsep=0pt, leftmargin=*]
    \setlength{\parskip}{0pt} 
    \item We propose the \textbf{phase manifold as a design prior} for cross-embodiment action embedding, where per-body-part parameters yield an interpretable and controllable action space. 
    \item We introduce a \textbf{human-anchored multi-embodiment alignment strategy} that extends a frozen phase oracle to four robots through lightweight adapters.
    \item Our structured embedding delivers \textbf{downstream gains} in motion imitation and teleoperation, with \textbf{strong cross-embodiment retrieval}.
\end{enumerate}
\vspace{-1pt}
\vspace{-2pt}
\section{Related Works}
\vspace{-1pt}
\textbf{Latent Action Representations for Robot Policy.} Modern robot policies learn an action representation that mediates between observation and motor command, formulating this space through autoregressive tokenization~\citep{zitkovich2023rt,kim2025openvla,team2024octo,bjorck2025gr00t}, diffusion and flow-matching denoising~\citep{chi2025diffusion,black2024pi_0}, or structured latents from video reconstruction~\citep{ye2025latent,bu2025univla,bu2025agibot} and discrete codebooks~\citep{mete2024quest,lee2024behavior,luo2025sonic}. 
These treat motion as an unstructured sequence, overlooking the periodic, morphology-invariant nature of humanoid movement. We instead build the action embedding around the periodic structure of motion.

\textbf{Periodic Representations in Motion.} Phase compactly describes where a behavior lies within a periodic cycle. Early methods modulated network weights from foot-contact heuristics~\citep{holden2017phase}, later extended to multi-contact and limb-specific phases~\citep{zhang2018mode,starke2019neural,starke2020local}, and unsupervised approaches learned multi-dimensional phase manifolds via FFT bottlenecks~\citep{starke2022deepphase}, extending to style modulation and in-betweening~\citep{mason2022real,zhang2025motion}. Recent work applies phase to human-quadruped alignment~\citep{li2024walkthedog}, functional autoencoders~\citep{pegoraro2025funphase}, per-body-part control~\citep{dai2026controllable}, and periodic skill discovery~\citep{park2026periodic}. Yet all remain confined to character animation, never used as action representations for humanoid robots or as a substrate for cross-embodiment alignment.

\textbf{Humanoid Cross-Embodiment Transfer.} Humanoid transfer can be grouped by alignment mechanism. Kinematic retargeting maps reference motion to a target body via inverse kinematics or interaction-aware optimization, yielding training data but no shared representation~\citep{gleicher1998retargetting,aberman2020skeleton,luo2023perpetual,luo2024universal,yang2025omniretarget}. Shared-latent approaches instead project motions into a common space via contrastive learning on extrinsic kinematics~\citep{yan2023imitationnet,cheng2024expressive}, scaled by per-part embeddings~\citep{yan2026learningunifiedlatentspace}, or unified through differentiable retargeting~\citep{qiu2025humanoidpolicyhuman}, visual transfer~\citep{wang2023mimicplaylonghorizonimitationlearning,dan2025x,kim2025uniskill}, and language-guided features~\citep{shao2025langwbc,yue2025rl,li2025language}. All rest on per-frame similarity over extrinsic or task-specific features, with no motion-intrinsic structure. We instead align heterogeneous robots on a structured, physically meaningful phase manifold.

\section{Preliminary: Periodic Motion Decomposition}
\label{sec:prelim}
\vspace{-0.5pt}


\textbf{Periodic Autoencoder (PAE).} 
Given a motion window $\mathbf{X} \in \mathbb{R}^{D \times T}$, a 1D convolutional encoder $g$ produces $M$ latent channels $\mathbf{L} = g(\mathbf{X}) \in \mathbb{R}^{M \times T}$, and an FFT-based bottleneck fits each channel to a single sinusoid with amplitude $A_c$, frequency $F_c$, phase shift $S_c$, and offset $B_c$, trained end-to-end under a reconstruction loss so that each channel captures one locally periodic component.
\begin{equation}
  \hat{l}_c(\tau) = A_c \cos\!\big(2\pi(F_c \tau + S_c)\big) + B_c
  \label{eq:pae-signal}
\end{equation}
\vspace{0.8pt}
\textbf{Phase Manifold.}
Each channel is projected onto an amplitude-scaled circle and stacked over channels to form a phase manifold $\mathbf{P}(\tau)=[m_1(\tau);\ldots;m_M(\tau)]\in\mathbb{R}^{2M}$
\vspace{0.2pt}
\begin{equation}
  \phi_c(\tau)=2\pi(\tau F_c + S_c), \qquad
  m_c(\tau)=A_c\big[\cos\phi_c(\tau),\;\sin\phi_c(\tau)\big]
  \label{eq:phase-embed}
\end{equation}
a counter-clockwise trajectory on a product of circles. Here $(A_c, F_c)$ encodes \emph{action identity} (e.g., walking vs.\ running) while $S_c$ encodes \emph{temporal position} within action, so points close on $\mathbf{P}$ are both semantically similar and temporally aligned~\cite{starke2022deepphase}.

\vspace{-0.2pt}
\begin{figure}[t]
\centering
\includegraphics[width=0.96\linewidth,height=0.2\textheight]{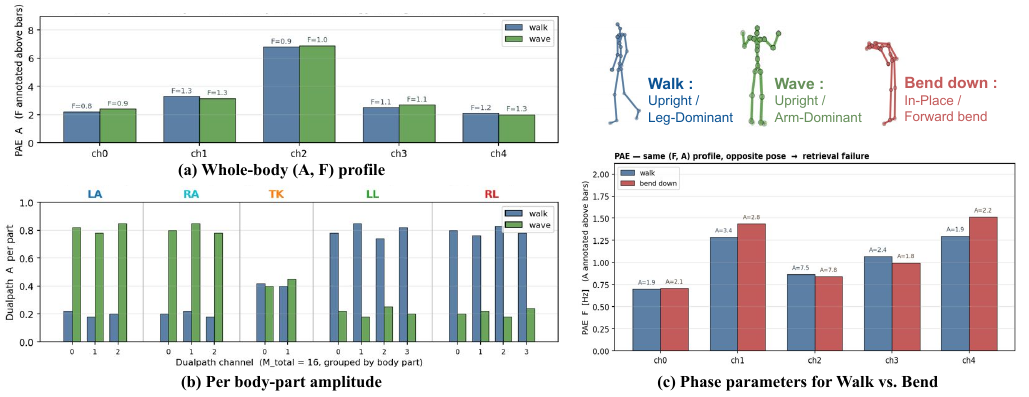}
\vspace{-0.2pt}
\caption{%
\textbf{Structural gaps of PAE.}
\textbf{(a,b)} Whole-body $(A, F)$ conflates distinct motions (walk vs.\ wave), which per-part features resolve. \textbf{(c)} Velocity-based $(A, F)$ ignores posture difference.
}
\label{fig:pae-limitations}
\end{figure}
\vspace{-0.2pt}

\textbf{Limitations for Cross-Embodiment Encoding.}
Applied to the cross-embodiment setting, this formulation has two structural gaps (\cref{fig:pae-limitations}). First, the encoder compresses the whole body into $M$ shared channels, so parts moving at independent rhythms (e.g., arms waving while walking) collapse into a weighted mixture (\cref{fig:pae-limitations}.(a)), erasing the part-level periodicity that is the natural unit of cross-embodiment correspondence (\cref{fig:pae-limitations}.(b)). Second, the encoding is velocity-based: it captures how the body moves but not where it is, so motions with similar $(A, F)$ but different posture map to nearby points (\cref{fig:pae-limitations}.(c)). We mitigate both gaps: \textbf{per-body-part encoding} recovers part-level periodicity, and \textbf{auxiliary pose branch} supplies static configuration without contaminating $\mathbf{P}(\tau)$.
\raggedbottom
\section{Method}
\begin{figure}[t]
\centering
\includegraphics[width=0.98\linewidth, height=0.35\textheight]{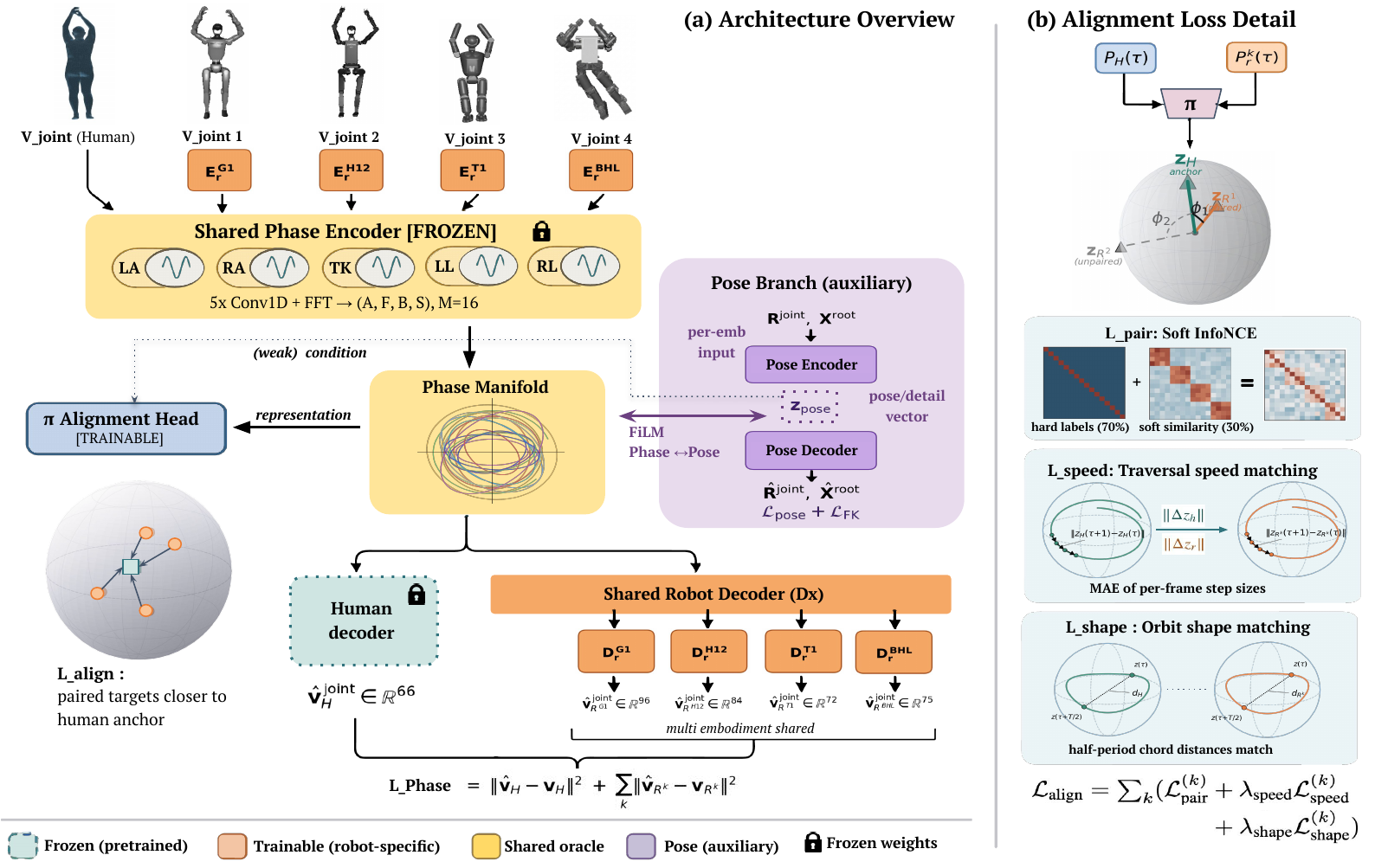}
\caption{\textbf{Overview of PHASOR.}
\textbf{(a)} A frozen human phase encoder, shared across embodiments via per-robot adapters, yields a morphology-invariant manifold $\mathbf{P}(\tau)$; a per-embodiment pose branch couples in only at the decoder via bidirectional FiLM.
\textbf{(b)} Manifolds are projected onto $\mathcal{S}^{63}$ and aligned by soft InfoNCE (LAMP prior) plus speed and shape consistency terms.}
\label{fig:main_figure_phasor}
\end{figure}

\vspace{-0.5pt}
We propose \textbf{PHASOR}, which learns a structured action embedding across one human and $K$ humanoid robots using periodic motion structure as its design prior. As shown in \cref{fig:main_figure_phasor}, motion is decomposed into a phase manifold $\mathbf{P}(\tau)$ that captures periodic dynamics and a complementary pose branch that captures non-periodic configuration. $\mathbf{P}(\tau)$ is the shared coordinate system for cross-embodiment alignment, while the pose branch stays per-embodiment. We first build this embedding for a single embodiment (\cref{sec:base_method}), then extend it across embodiments via a human-anchored scheme (\cref{sec:cross}), training in two stages: establish the human oracle, then align robots on top of it.

\subsection{From Periodic Prior to Structured Action Manifold}
\label{sec:base_method}

\textbf{Task Definition.}
We consider one human ($J_H{=}22$ SMPL-H joints) and $K$ robots with joint counts $J_k$. Each motion is a root-local window of $T{=}121$ frames ($\approx$2\,s, 60\,fps), with joint velocity $\mathbf{v}^{\text{joint}}\in\mathbb{R}^{T\times 3J}$, 6D joint rotation $\mathbf{R}^{\text{joint}}\in\mathbb{R}^{T\times 6J}$, and root position $\mathbf{X}^{\text{root}}\in\mathbb{R}^{T\times 3}$, on normalized time $\tau\in[-\tfrac{T}{2\,\mathrm{fps}},\tfrac{T}{2\,\mathrm{fps}}]$. Our goal is a phase manifold $\mathbf{P}(\tau)\in\mathbb{R}^{T\times 2M}$, paired with a complementary pose code for non-periodic posture, such that semantically equivalent windows from different embodiments map to nearby trajectories on a shared coordinate system while staying interpretable and controllable per embodiment.

\subsubsection{Per-Part Periodic Decomposition}
\label{sec:perpart}
A whole-body encoder conflates joints with independent rhythms (\cref{sec:prelim}). We instead partition every embodiment into five kinematic parts $\mathcal{B}=\{\text{LA},\text{RA},\text{TK},\text{LL},\text{RL}\}$ (left/right arm, trunk, left/right leg), each with its own phase encoder $\Phi_b$ and periodic bottleneck; per-embodiment joint-to-part assignments follow kinematic role (\cref{app:body_parts}). For each part $b$, $\Phi_b$ models the velocity signal $\mathbf{v}_b(\tau)$ as $K_b$ sinusoids, extracting $(A_{b,c}, F_{b,c}, B_{b,c})$ via a differentiable FFT layer and $S_{b,c}$ via a regression head~\cite{starke2022deepphase}. Channels are allocated by each part's rhythmic complexity, $M=16$ in total.
\begin{equation}
  \mathbf{v}_b(\tau)\approx\sum_{c=1}^{K_b} A_{b,c}\cos\!\big(2\pi(F_{b,c}\tau + S_{b,c})\big) + B_{b,c},
  \label{eq:per_part_signal}
\end{equation}

We then project every channel onto its amplitude-scaled circle via \cref{eq:phase-embed}, now indexed by part as $m_{b,c}(\tau)=A_{b,c}[\cos\phi_{b,c}(\tau),\sin\phi_{b,c}(\tau)]$, and concatenate over all $(b,c)$ to form the part-structured manifold $\mathbf{P}(\tau)\in\mathbb{R}^{T\times 32}$; a symmetric per-part decoder $D_b$ reconstructs \cref{eq:per_part_signal}. Independent encoders $\Phi_b$ keep each part's rhythmic signature free of cross-part contamination.



\subsubsection{Complementary Pose Conditioning}
\label{sec:method_pose}
Velocity-based phase lacks static postural context (e.g., crouching vs.\ standing), so we add a complementary pose branch that encodes 6D joint rotations and root positions into a compact token sequence $\mathbf{z}_{\text{pose}}\in\mathbb{R}^{L_p\times d_p}$ ($L_p{=}8$, $d_p{=}128$), retaining only window-level postural configuration. The two branches interact exclusively at decoding via bidirectional FiLM~\cite{perez2018film}: phase context conditions pose decoding, while pose
context modulates the reconstructed velocity signal before the phase decoder. Crucially, $\mathbf{P}(\tau)$ is constructed from FFT parameters \emph{before} any modulation, keeping the manifold pose-free and suitable as a clean cross-embodiment alignment target (\cref{sec:cross}).

\subsection{Cross-Embodiment Alignment on the Phase Manifold}
\label{sec:cross}
\vspace{-0.5pt}
As the channels of $\mathbf{P}(\tau)$ carry rhythmic identity rather than morphology, we treat it as a shared coordinate system. Every robot is mapped onto the \textbf{\emph{same}} manifold and aligned by semantic correspondence. We realize this by reusing the human-pretrained backbone as a frozen oracle and routing each robot through it with lightweight adapters.

\subsubsection{Human-Anchored Encoder}
\label{sec:anchor}
Human data is abundant and behaviorally diverse while robot data is sparse, so we pretrain the backbone on human motion and freeze it as a \textbf{shared anchor} where every robot is aligned to. The per-part encoders $E_h=\{\Phi_b\}_{b\in\mathcal{B}}$ and the phase-to-pose FiLM head are frozen, fixing the channel semantics once so new robots join without representation drift. For robot $k$, a linear input adapter $E_r^k:\mathbb{R}^{D_k}\to\mathbb{R}^{66}$ maps its joint velocities into the human input space, and the shared frozen encoder $E_X\equiv E_h$ extracts the phase parameters:
\begin{equation}
  \mathbf{v}^{\text{joint}}_{R^k}\xrightarrow{E_r^k}\tilde{\mathbf{v}}^{(66)}_{R^k}\xrightarrow{E_X\equiv E_h}(A_{b,c}, F_{b,c}, B_{b,c}, S_{b,c})\xrightarrow{\text{Eq.~\ref{eq:per_part_signal}}}\mathbf{P}_{R^k}(\tau).
  \label{eq:robot-pipeline}
\end{equation}

Robot joints use the same five-part partition (\cref{app:body_parts}). Since $E_X$ is shared, all morphological variation is absorbed by the small adapter $E_r^k$, and $\mathbf{P}_{R^k}(\tau)$ adheres to the human phase prior.


\subsubsection{Decoder Adaptation and Pose Isolation}
\label{sec:robot_pipeline}
The decoder mirrors the encoder but allows controlled adaptation: the human phase decoder $D_h$ stays frozen, while robots share a phase decoder $D_X$ initialized from $D_h$ and fine-tuned with per-robot output adapters $D_r^k:\mathbb{R}^{66}\to\mathbb{R}^{D_k}$. Because posture is morphology-dependent, each robot has its own pose branch supervised by reconstruction only, keeping morphology-specific detail out of the shared-manifold alignment.

\subsubsection{Alignment Objective}
\label{sec:align_obj}
All supervision acts on $\mathbf{P}(\tau)$, not on the pose branch, keeping the signal independent of kinematics.

\textbf{Alignment head.}
For latent alignment, we concatenate the phase manifold $\mathbf{P}(\tau)$ with a temporally pooled pose token $\mathbf{z}^{\text{pool}}_{\text{pose}}$ and project the result onto a \textbf{unit sphere} via a lightweight head $\pi$.
\begin{equation}
  z(\tau) = \pi\big([\,\mathbf{P}(\tau)\,;\, \mathbf{z}^{\text{pool}}_{\text{pose}}\,]\big) \in \mathcal{S}^{63}.
  \label{eq:align-projection}
\end{equation}
The pose token is added to disambiguate motions that share the same rhythm but hold a different posture, which phase alone cannot tell apart. To keep alignment driven mainly by phase, the pose-related weights in $\pi$ are initialized near zero, so the head starts phase-only and brings in pose context gradually. Time-pooling $z(\tau)$ over a clip gives a clip-level descriptor $z^{\text{pool}}\in\mathcal{S}^{63}$.

\textbf{Hierarchical Pair Loss.} For each human-robot pair we use an InfoNCE objective whose target mixes the hard diagonal label with a soft target from a motion-semantic prior (LAMP~\cite{li2025lamp}).
\begin{equation}
  T^{\text{mix}} = (1 - \lambda_{\text{mix}}) T^{\text{hard}} + \lambda_{\text{mix}} T^{\text{soft}}_{\text{LAMP}}
  \label{eq:soft-target}
\end{equation}
The hard term anchors paired windows, while the LAMP term pulls together semantically similar motions that hard labels treat as negatives. We apply $T^{\text{mix}}$ at two granularities, per-window for \textbf{\textit{fine}} motion variation and per-clip (mean-pooled $z$ within a clip) for \textbf{\textit{coarse}} meaning, summing the symmetric cross-entropy of both into $\mathcal{L}_{\text{pair}}$.

\textbf{Trajectory Consistency.} Two lightweight geometric penalties regularize the temporal shape of $z(\tau)$: $\mathcal{L}_{\text{speed}}$ matches per-frame step size on the sphere (implicit frequency synchronization) and $\mathcal{L}_{\text{shape}}$ matches half-period distance (implicit amplitude synchronization). The full alignment loss is
$\mathcal{L}_{\text{align}} = \sum_k\big(\mathcal{L}^{(k)}_{\text{pair}} + \lambda_{\text{speed}}\mathcal{L}^{(k)}_{\text{speed}} + \lambda_{\text{shape}}\mathcal{L}^{(k)}_{\text{shape}}\big)$.


\subsubsection{Training}
\label{sec:training}
We optimize the framework in two continuous phases. We first pretrain the human backbone with a per-part velocity reconstruction loss $\mathcal{L}_{\text{phase}}$, a pose reconstruction loss $\mathcal{L}_{\text{pose}}$, and a forward-kinematics consistency loss $\mathcal{L}_{\text{FK}}$ that geometrically links the two branches. We then freeze the human oracle and jointly train the robot components (input/output adapters, shared phase decoder $D_X$, per-robot pose branches, and alignment head $\pi$) under \cref{eq:total_loss}. 
\begin{equation}
    \mathcal{L}_{\text{total}} = \lambda_{\text{phase}} \mathcal{L}_{\text{phase}} + \lambda_{\text{pose}} \mathcal{L}_{\text{pose}} + \lambda_{\text{FK}} \mathcal{L}_{\text{FK}} + \mathcal{L}_{\text{align}}
    \label{eq:total_loss}
\end{equation}
A short reconstruction-only warm-up precedes alignment, so each $\mathbf{P}_{R^k}$ stabilizes before contrastive supervision begins. Full details are provided in \cref{app:training}.

\begin{table*}[!t]
\centering
\setlength{\tabcolsep}{3pt}
\renewcommand{\arraystretch}{1.0}
\small
\resizebox{\linewidth}{!}{%
\begin{tabular}{l | cccc | cccc | cccc}
\toprule
 & \multicolumn{4}{c|}{\textbf{H$\to$R}} & \multicolumn{4}{c|}{\textbf{R$\to$H}} & \multicolumn{4}{c}{\textbf{R$\to$R}} \\
\cmidrule(lr){2-5}\cmidrule(lr){6-9}\cmidrule(lr){10-13}
Method & R@1 & R@5 & R@10 & MRR & R@1 & R@5 & R@10 & MRR & R@1 & R@5 & R@10 & MRR \\
\midrule
MLP~\citep{yan2026learningunifiedlatentspace,yan2023imitationnet}                                           & 79.9 & 96.5 & 98.8 & 87.1 & 79.8 & \underline{97.5} & \underline{99.3} & 87.4 & 74.7 & \underline{95.0} & \underline{98.2} & \underline{83.5} \\
VQ~\citep{ye2025latent,luo2025sonic}                        & 37.6 & 78.6 & 91.0 & 55.1 & 36.3 & 77.0 & 89.5 & 53.6 & 37.5 & 77.7 & 89.8 & 54.8 \\
\midrule
\rowcolor{black!10}
\multicolumn{13}{l}{\emph{(A) Alignment loss design}} \\
\midrule
Only Phase (zero-shot)                         & 83.0 & 91.9 & 93.6 & 87.0 & 83.9 & 92.2 & 94.1 & 87.8 & 73.0 & 84.3 & 87.3 & 78.2 \\
Hard only (no LAMP)                            & 84.1 & 92.2 & 93.8 & 87.8 & 84.1 & 92.5 & 94.2 & 87.8 & 74.5 & 85.3 & 88.1 & 79.5 \\
\;$+$ Soft Coarse                              & 84.9 & 92.7 & 94.3 & 88.5 & 85.1 & 93.0 & 94.6 & 88.6 & 76.1 & 86.4 & 88.9 & 80.7 \\
\;$+$ Soft Fine                                & 85.6 & 93.4 & 94.9 & 89.1 & 85.7 & 93.7 & 95.2 & 89.1 & 77.2 & 87.1 & 88.6 & 81.6 \\
\;$+$ Soft Coarse$+$Fine                       & 86.9 & 94.3 & 95.8 & 89.9 & 87.2 & 94.8 & 96.3 & 90.2 & 79.3 & 89.6 & 91.8 & 83.7 \\
\midrule
\rowcolor{black!10}
\multicolumn{13}{l}{\emph{(B) Alignment target} \,(loss $=$ Soft Coarse$+$Fine)} \\
\midrule
$\pi(\mathbf{P}_\tau \,\|\, \mathbf{X}_\text{root})$                          & \underline{88.5} & \underline{96.0} & \underline{97.2} & \underline{91.9} & \underline{89.5} & 96.5 & 97.5 & \underline{92.6} & \underline{81.0} & 91.7 & 93.9 & 85.8 \\
\rowcolor{blue!10}
$\pi(\mathbf{P}_\tau \,\|\, \mathbf{z}_\text{pose})$ \textbf{(Ours)}          & \textbf{90.3} & \textbf{98.3} & \textbf{99.2} & \textbf{93.9} & \textbf{90.5} & \textbf{98.6} & \textbf{99.4} & \textbf{94.1} & \textbf{84.8} & \textbf{96.4} & \textbf{98.3} & \textbf{90.0} \\
\bottomrule
\end{tabular}}
\caption{\textbf{Cross-embodiment retrieval and PHASOR ablation.} Each query has one positive, the temporally identical chunk from another embodiment (R$\to$R over $\binom{4}{2}{=}6$ pairs, \%). \textbf{(A)} varies the loss at a fixed target, adding LAMP soft targets at the \textbf{\textit{coarse}} clip, \textbf{\textit{fine}} window, and \textbf{\textit{both}} levels. \textbf{(B)} varies the head input at the best loss, $\mathbf{X}_\text{root}$ vs.\ our pooled $\mathbf{z}_\text{pose}$. \textbf{Bold} best, \underline{underline} second.}
\label{tab:retrieval_main}
\end{table*}
\vspace{-2pt}

\section{Experiments}
\label{sec:experiments}

We evaluate PHASOR along two axes. \textbf{\emph{Embedding quality}} asks whether the phase manifold yields an interpretable, cross-embodiment-aligned action space (\cref{sec:exp_embedding}), and \textbf{\emph{downstream utility}} asks whether that embedding produces concrete gains once frozen and plugged into robot policy learning (\cref{sec:exp_downstream}). In every downstream evaluation the phase encoder stays frozen and only a lightweight policy head is trained, so any performance gap comes from representation quality alone.
 
\textbf{Data.} We use 12,616 clips (${\sim}$40.4\,h, 8.7\,M frames at 60\,fps) from AMASS CMU and KIT, each retargeted with GMR to four humanoids, Unitree G1 (32 DOF), H1.2 (28 DOF), Booster T1 (24 DOF), and Berkeley Humanoid Lite (25 DOF), yielding 50,464 synchronized human-robot pairs. Every experiment uses a 121-frame chunk (${\sim}$2\,s).

\textbf{Baselines.} We compare two cross-embodiment embedding strategies. \textbf{MLP}~\citep{yan2026learningunifiedlatentspace,yan2023imitationnet} projects per-embodiment features into 80D with a segment-wise MLP under InfoNCE, a contrastive baseline with no structural prior. \textbf{VQ}~\citep{ye2025latent,luo2025sonic} encodes motion into a 10D discrete codebook via cross-reconstruction, isolating discrete quantization against our periodic prior. As objectives vary across quantizer-based methods, we use a representative codebook variant so the comparison reflects quantization itself.

\subsection{Embedding Quality}
\label{sec:exp_embedding}

\subsubsection{Cross-Embodiment Retrieval}
\label{sec:exp_retrieval}

We measure alignment with chunk retrieval. From 100 held-out clips we sample 20 non-overlapping 121-frame chunks each and query the gallery of a different embodiment, where each query has one positive, the temporally identical chunk. We report Recall@$k$ and Mean Reciprocal Rank (MRR) over three directions, Human$\to$Robot (H$\to$R), Robot$\to$Human (R$\to$H), and Robot$\to$Robot (R$\to$R, averaged over all $\binom{4}{2}{=}6$ pairs).

\textbf{Baselines.} MLP reaches reasonable exact-match scores since its objective directly optimizes per-frame discrimination, whereas VQ performs far worse, as its codebook maps adjacent chunks to the same code and collapses the fine-grained distinctions retrieval needs.

\textbf{(A) Alignment loss.} Applied to robots without cross-embodiment training, the frozen human-only manifold (\emph{Only Phase}) already beats MLP on H$\to$R (83.0 vs.\ 79.9 R@1), confirming a strong morphology-invariant temporal prior. Cross-embodiment InfoNCE (\emph{Hard only}) improves every direction, and LAMP soft targets help further, where coarse clip-level (\emph{Soft Coarse}) and fine window-level (\emph{Soft Fine}) targets each raise accuracy and their combination (\emph{Soft Coarse$+$Fine}) is best in (A) at 86.9 R@1. The R$\to$R gain is most telling, as robot pairs are never directly aligned in training. Alignment emerges transitively through the shared human anchor, which the LAMP prior, common to all embodiments, makes possible.

\textbf{(B) Alignment target.} With the loss fixed to Soft Coarse$+$Fine, we change what the head receives alongside $\mathbf{P}(\tau)$. Adding the raw root position $\mathbf{X}_\text{root}$ lifts R@1 to 88.5 on H$\to$R and our pooled, FiLM-modulated pose token $\mathbf{z}_\text{pose}$ (\emph{Ours}) is best on every metric and direction (90.3 R@1, 93.9 MRR on H$\to$R). The modulated token adds postural context already consistent with the phase manifold, separating same-rhythm motions without injecting raw morphology.

\subsubsection{Manifold Visualization and Interpretability}
\label{sec:exp_interp}
 

\begin{figure*}[t]
\centering
\begin{minipage}[t]{0.49\linewidth}
  \centering
    \includegraphics[width=\linewidth,height=0.21\textheight,keepaspectratio]{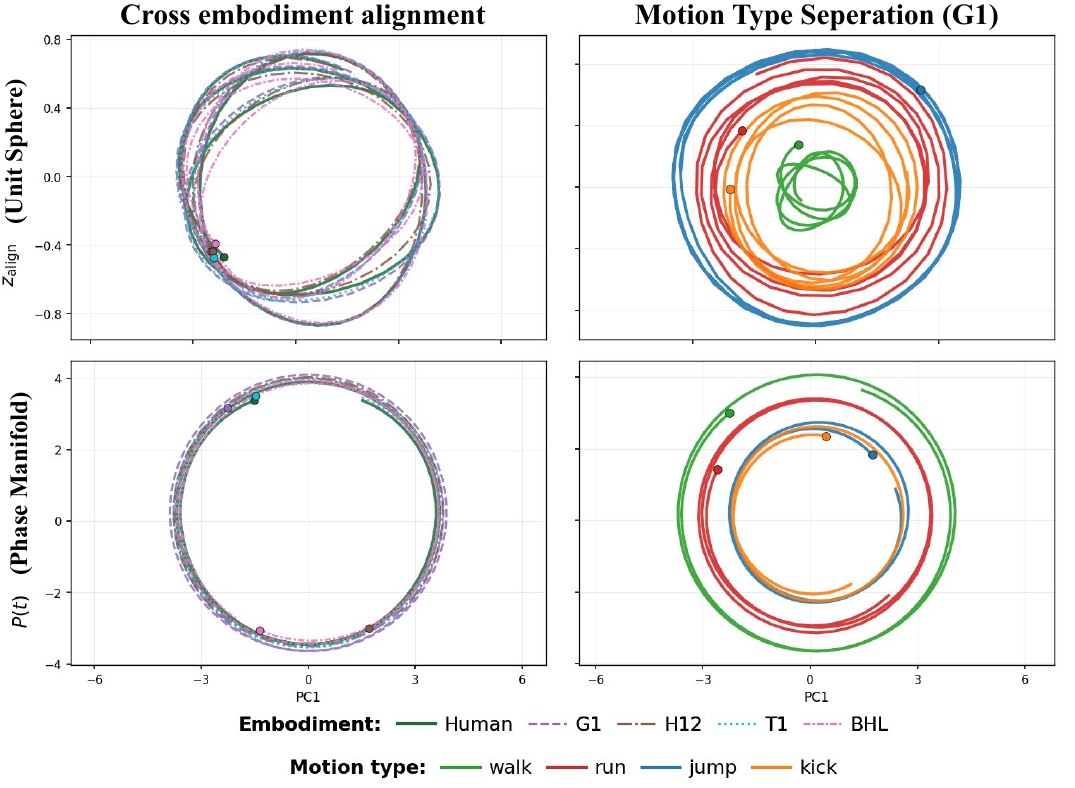}
\end{minipage}
\hfill
\begin{minipage}[t]{0.5\linewidth}
  \centering
    \includegraphics[width=\linewidth,height=0.21\textheight,keepaspectratio]{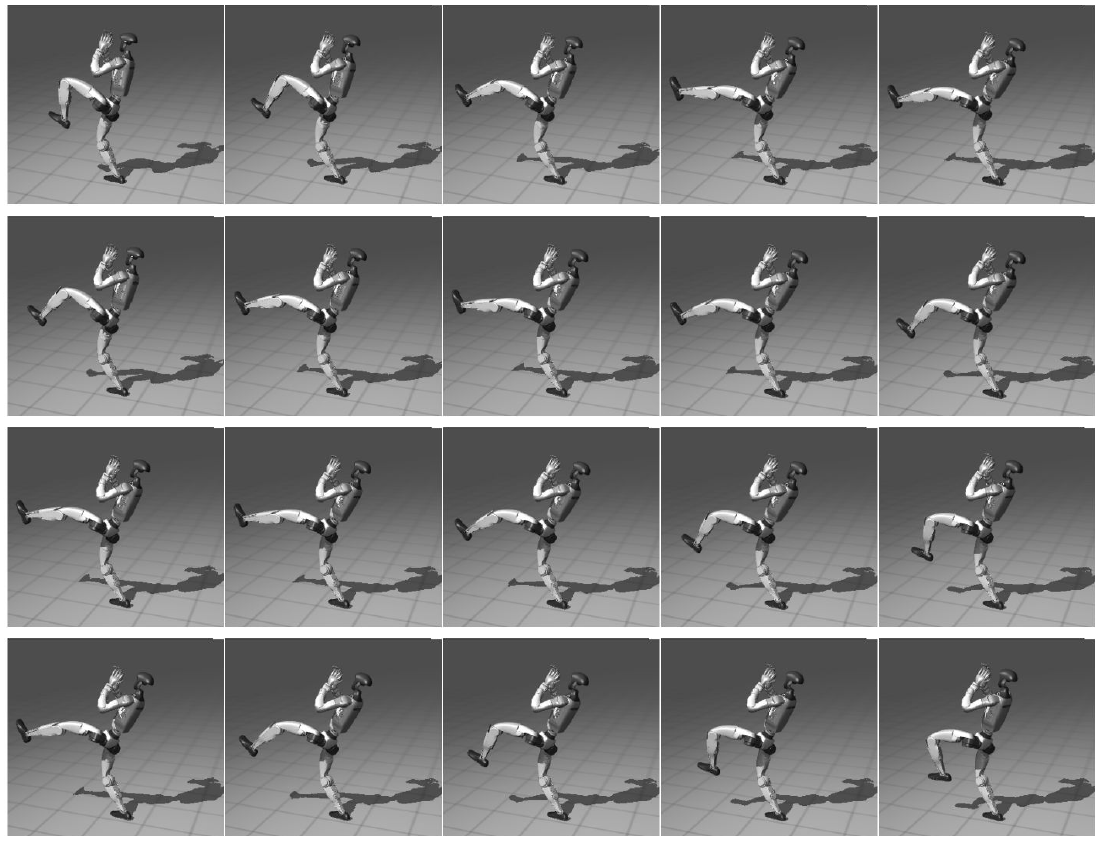}
\end{minipage}
\caption{\textbf{Left.} 2D PCA of the alignment embedding $z$ (top) and phase manifold $\mathbf{P}(\tau)$ (bottom). Same clip across the human and four robots overlaps tightly (col.\ 1), while G1 separates four motion types (walk, run, jump, kick) by orbit radius and position (col.\ 2). \textbf{Right.} Phase shifts $\Delta S$ on a running clip yield predictable temporal offsets, each row at a different gait phase.}
\label{fig:manifold_vis}
\end{figure*}

The PCA projections (left of \cref{fig:manifold_vis}) reveal two properties. \emph{Intra-clip continuity} (second column), within a clip the advancing phase $\phi_c(\tau)$ traces a smooth counter-clockwise orbit whose radius reflects amplitude, a signature unstructured embeddings lack. \emph{Cross-embodiment overlap} (first column), matched timestamps from different embodiments cluster tightly, validating the human-anchored alignment. The alignment embedding $z$ (top) is mildly distorted relative to the raw manifold $\mathbf{P}(\tau)$ (bottom) yet keeps the circular structure and still separates motion types.

We further show that the phase shift $S_c$ is an interpretable, controllable temporal coordinate (right of \cref{fig:manifold_vis}). For a G1 running clip we read per-window $S_c$ with the frozen encoder, unwrap it into a monotone timeline, and retrieve windows at evenly spaced offsets $S_q + \Delta S$. The retrieved start point advances in proportion to $\Delta S$, confirming the embedding captures motion timing and supports per-embodiment phase control.
\subsection{Downstream Tasks}
\label{sec:exp_downstream}
\begin{figure}[t]
\centering
\begin{minipage}[c]{0.20\linewidth}
  \centering
  \footnotesize
  \setlength{\tabcolsep}{4pt}   
  \begin{tabular}{lcccc}
    \toprule
    & \multicolumn{2}{c}{\textbf{Imitation}} 
    & \multicolumn{2}{c}{\textbf{Teleoperation}} \\
    \cmidrule(lr){2-3}\cmidrule(lr){4-5}
    \textbf{Variant} & $d_z$ & MPJPE$\downarrow$ 
                     & $d_z$ & MPJPE$\downarrow$ \\
    \midrule
    Propr. raw & -- & 1.83 & -- & 64.85 \\
    MLP        & 80 & 1.83 & 80 & 66.79 \\
    VQ         & 10 & 1.69 & 10 & 65.17 \\
    \textbf{PHASOR} & 64 & \textbf{1.62} & 64 & \textbf{64.75} \\
    \bottomrule
  \end{tabular}
\end{minipage}
\hfill
\begin{minipage}[c]{0.57\linewidth}
  \centering
  \includegraphics[width=\linewidth]{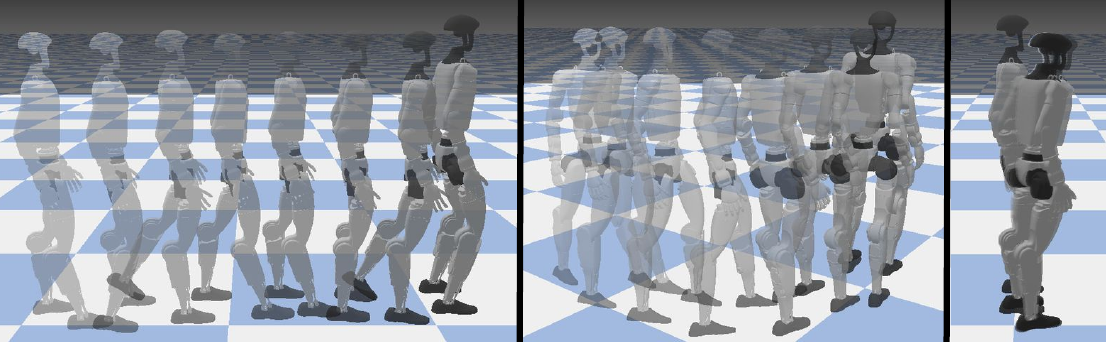}
\end{minipage}

\caption{\textbf{Left.} Downstream results, all encoders frozen and only the policy head trained (MPJPE in mm, $\downarrow$). Propr.\ raw uses raw proprioceptive input with no embedding. \textbf{Right.} IsaacSim walking-forward rollout, left to right PHASOR (ours), VQ, MLP.}
\label{fig:downstream}
\end{figure}

We test the frozen embedding on three tasks, each freezing the encoder and training only a lightweight policy head to isolate representation quality. As a floor, we add a \emph{Proprioceptive raw} baseline that drives the policy head from raw proprioceptive input alone, with no learned embedding (first row of \cref{fig:downstream}). The exact proprioceptive inputs and downstream settings are in \cref{app:rl_reward}.

\textbf{Motion Imitation and Teleoperation.} Both predict 3D joint positions from a frozen embedding via a 3-layer MLP. In \emph{imitation}, the G1's proprioceptive state and $z$ predict its next-frame pose, where PHASOR scores best (\cref{fig:downstream}) as its explicit phase parameterization pins down the gait-cycle position while the VQ codebook captures periodicity only implicitly. In \emph{teleoperation}, a human window maps to the matching G1 pose from $z$ alone, where only PHASOR beats the raw-kinematics baseline and MLP falls below it, showing an unstructured projection can discard useful information.


\textbf{RL Locomotion with Phase Reward.} Here the embedding shapes the reward rather than the policy input. The same frozen encoder $f$ embeds the robot's online motion and the human reference, and the reward scores their alignment,
\begin{equation}
  r_\text{phase} = \lambda_\text{phase}\,\big(\cos(f(x_\text{robot}),\, f(x_\text{ref})) + 1\big)/2,
  \label{eq:phase-reward}
\end{equation}
so each method measures phase agreement in \emph{its own} space without prescribing a joint-level solution. The reference enters only through the reward and all variants share identical settings, isolating embedding geometry. In \cref{fig:downstream}, PHASOR walks forward steadily on both feet, VQ under-uses one leg and drifts in a circle, and MLP fails to locomote, confirming our phase embedding gives an effective reward for periodic, walking-like motion.

\section{Conclusion}
\label{sec:conclusion}
We present PHASOR, an action representation that treats the periodic structure of motion as a design prior rather than learning the latent space from reconstruction alone. By decomposing motion into per-body-part phase parameters, PHASOR builds a compact, morphology-invariant manifold anchored in human motion, where each new robot joins through a lightweight adapter against a frozen oracle, yielding an interpretable and controllable action space whose coordinates expose behavior timing. Across one human and four humanoids, this frozen manifold transfers effectively to retrieval and downstream policy learning, indicating that motion-intrinsic periodicity is a transferable substrate for cross-embodiment action embeddings.

\textbf{Limitations.} PHASOR is built around periodic structure, so it is most informative for rhythmic, locomotion-style behaviors while providing weaker structure for transient or aperiodic motions, and its pose branch remains per-embodiment, leaving non-periodic posture unshared across bodies. Validation is currently limited to simulation, with physical hardware left to future work. We view these as natural next steps rather than fundamental barriers.


	


\clearpage


\bibliography{main}  

\clearpage
\newpage
\onecolumn
\appendix

\pagenumbering{roman}
\renewcommand\thetable{\Roman{table}}
\renewcommand\thefigure{\Roman{figure}}
\setcounter{section}{0}
\setcounter{table}{0}
\setcounter{figure}{0}

\noindent\textbf{\Large Appendix}

\section{Implementation Details}

\subsection{Body-Part Partition and Channel Allocation}
\label{app:body_parts}

To systematically capture local periodic structures, we partition kinematic joints of each embodiment into five semantic parts: $\mathcal{B} = \{\text{LA}, \text{RA}, \text{TK}, \text{LL}, \text{RL}\}$. We assign the phase channel allocation $K_b$ based on the rhythmic complexity of each part during locomotion. Specifically, legs exhibit strong multi-phase oscillations requiring more capacity ($K_{\text{LL}}=K_{\text{RL}}=4$), arms show moderate complexity ($K_{\text{LA}}=K_{\text{RA}}=3$), and the trunk's dominant modes are set to $K_{\text{TK}}=2$. 

This allocation is shared across all embodiments, yielding a universal $M_{\text{total}} = 16$ channels and a 32-dimensional phase manifold $\mathbf{P}(\tau)$. \cref{tab:body-parts} summarizes the joint distributions and kinematic dimensions for the human and four humanoid robots.

\begin{table}[h]
\centering
\small
\begin{tabular}{l cc | ccccc | l}
\toprule
\multirow{2}{*}{\textbf{Embodiment}} & \multirow{2}{*}{$\boldsymbol{J}$} & \multirow{2}{*}{$\boldsymbol{D_k}$} & \multicolumn{5}{c|}{\textbf{Joints per part}} & \multirow{2}{*}{\textbf{TK Components}} \\
 & & & LA & RA & TK & LL & RL & \\
\midrule
Human (SMPL-H) & 22 & 66 & 4 & 4 & 6 & 4 & 4 & pelvis, spine (3), neck, head \\
G1 (Unitree)   & 32 & 96 & 7 & 7 & 4 & 7 & 7 & pelvis, waist (2), torso \\
H12 (Unitree)  & 28 & 84 & 7 & 7 & 2 & 6 & 6 & pelvis, torso \\
T1 (Booster)   & 24 & 72 & 4 & 4 & 4 & 6 & 6 & trunk, head (2), waist \\
BHL (Berkeley) & 25 & 75 & 6 & 6 & 1 & 6 & 6 & base only \\
\midrule
\multicolumn{3}{l|}{\textbf{Phase channels ($K_b$)}} & \textbf{3} & \textbf{3} & \textbf{2} & \textbf{4} & \textbf{4} & $M_{\text{total}} = 16$ \\
\bottomrule
\end{tabular}
\vspace{4pt}
\caption{\textbf{Body-part partition and channel allocation.} We enforce a consistent five-part semantic structure and phase channel allocation ($K_b$) across heterogeneous embodiments. $J$ denotes total joints, and $D_k$ is velocity dimension ($3J$). The 3D root linear velocity is concatenated to the trunk (TK) input.}
\label{tab:body-parts}
\end{table}
\begin{table}[h]
\centering
\small
\resizebox{\columnwidth}{!}{%
\begin{tabular}{l l p{0.65\linewidth}}
\toprule
\textbf{Embodiment} & \textbf{Part} & \textbf{Joint Indices \& Description} \\
\midrule
\multirow{3}{*}{\shortstack[l]{Human\\(SMPL-H)}} 
& LA / RA & [13, 16, 18, 20] / [14, 17, 19, 21]: collar, shoulder, elbow, wrist \\
& TK      & [0, 3, 6, 9, 12, 15]: pelvis, spine 1--3, neck, head \\
& LL / RL & [1, 4, 7, 10] / [2, 5, 8, 11]: hip, knee, ankle, foot \\
\midrule
\multirow{3}{*}{\shortstack[l]{G1\\(Unitree)}} 
& LA / RA & [18--24] / [25--31]: shoulder (3), elbow, wrist (3) \\
& TK      & [0, 15--17]: pelvis, waist (2), torso \\
& LL / RL & [1--7] / [8--14]: hip (3), knee, ankle (2), toe \\
\midrule
\multirow{3}{*}{\shortstack[l]{H12\\(Unitree)}} 
& LA / RA & [14--20] / [21--27]: shoulder (3), elbow, wrist (3) \\
& TK      & [0, 13]: pelvis, torso \\
& LL / RL & [1--6] / [7--12]: hip (3), knee, ankle (2) \\
\midrule
\multirow{3}{*}{\shortstack[l]{T1\\(Booster)}} 
& LA / RA & [3--6] / [7--10]: arm joints 1--3, hand \\
& TK      & [0--2, 11]: trunk, head 1--2, waist \\
& LL / RL & [12--17] / [18--23]: hip (3), shank, ankle (cross), foot \\
\midrule
\multirow{3}{*}{\shortstack[l]{BHL\\(Berkeley)}} 
& LA / RA & [1--6] / [7--12]: shoulder (3), elbow (2), hand \\
& TK      & [0]: base only \\
& LL / RL & [13--18] / [19--24]: hip (3), knee, ankle (2) \\
\bottomrule
\end{tabular}
}
\vspace{4pt}
\caption{\textbf{Explicit joint index mappings.} Joints are assigned to the five semantic parts following each embodiment's specific URDF hierarchy. Left/Right symmetric parts are grouped for brevity, and the numbers in parentheses indicate the degrees of freedom (e.g., pitch, roll, yaw) for that joint group.}
\label{tab:joint-mapping}
\end{table}

\subsection{Bidirectional FiLM Coupling Architecture}
\label{app:film_architecture}

We provide the full architectural details of the bidirectional FiLM coupling described in \cref{sec:method_pose}. Both directions pass through a learned manifold encoder ($\mathcal{E}_\psi$ for phase-to-pose, $\mathcal{E}_\phi$ for pose-to-phase) that aligns the source representation to the temporal granularity of the target branch.

\textbf{Phase-to-Pose Modulation.}
The phase-to-pose direction provides the pose decoder with temporal context: ``where in the action the body currently is''. The manifold encoder $\mathcal{E}_\psi$ compresses the per-frame phase manifold $\mathbf{P} \in \mathbb{R}^{T \times 32}$ into a token-aligned context $\mathcal{E}_\psi(\mathbf{P}) \in \mathbb{R}^{L_{\text{pose}} \times d_g}$ with $d_g = 128$, implemented as three 1D convolutions followed by adaptive pooling to $L_{\text{pose}} = 8$ tokens. A FiLM head produces per-token modulation parameters $(\gamma, \beta) \in \mathbb{R}^{L_{\text{pose}} \times d_{\text{pose}}}$ that scale and shift the pose tokens:
\begin{equation}
  \mathbf{z}_{\text{pose}}^{\text{mod}} = (1 + \alpha\gamma) \odot \mathbf{z}_{\text{pose}} + \alpha\beta.
  \label{eq:phase-to-pose-film}
\end{equation}

\textbf{Pose-to-Phase Modulation.}
Symmetrically, the pose-to-phase direction provides each per-part decoder with postural context: ``what baseline configuration the motion rests on''. The manifold encoder $\mathcal{E}_\phi$ expands the pose tokens $\mathbf{z}_{\text{pose}} \in \mathbb{R}^{L_{\text{pose}} \times d_{\text{pose}}}$ into per-frame context $\mathcal{E}_\phi(\mathbf{z}_{\text{pose}}) \in \mathbb{R}^{T \times d_{g,\text{pose}}}$ with $d_{g,\text{pose}} = 64$, implemented as two 1D convolutions followed by linear interpolation from $L_{\text{pose}} = 8$ tokens to $T = 121$ frames. A per-part FiLM head generates per-frame modulation parameters $(\gamma_b(\tau), \beta_b(\tau)) \in \mathbb{R}^{K_b}$ that modulate the reconstructed sinusoidal signal $\hat{\mathbf{v}}_b^{\text{sig}}(\tau) = \sum_k A_{b,k}\cos\phi_{b,k}(\tau) + B_{b,k}$ before it enters the per-part decoder $D_b$:
\begin{equation}
  \hat{\mathbf{v}}_b^{\text{sig,mod}}(\tau) = \big(1 + \alpha\gamma_b(\tau)\big) \odot \hat{\mathbf{v}}_b^{\text{sig}}(\tau) + \alpha\beta_b(\tau).
  \label{eq:pose-to-phase-film}
\end{equation}

\textbf{Modulation Strength.}
The scalar $\alpha$ is learnable in both directions and initialized to $0.05$. This near-zero start ensures that the modulation begins close to identity, providing the same training stability while allowing the model to gradually scale up modulation as training progresses.


\subsection{Periodic Module Details}
\label{app:phase_module}

We detail the per-part periodic encoder $\Phi_b$, the FFT-based parameter extraction, and the handling of root velocities.

\textbf{Per-Part Encoder and Decoder.}
Each $\Phi_b$ is a lightweight 1D convolutional encoder of two layers with LayerNorm over the time dimension and ELU activations. The latent map $\Phi_b(\mathbf{v}_b) \in \mathbb{R}^{K_b \times T}$ feeds the FFT bottleneck. The per-part decoder $D_b$ mirrors $\Phi_b$ with the same width and depth, taking the modulated sinusoidal signal $\hat{\mathbf{v}}_b^{\text{sig,mod}}(\tau)$ and producing the reconstructed velocity. For the trunk, the decoder output is sliced into the joint-velocity ($\hat{\mathbf{v}}^{\text{joint}}_{\text{TK}}$) and the root linear velocity portion ($\hat{\mathbf{v}}^{\text{root}}_{\text{lin}}$).

\textbf{FFT Parameter Extraction.}
Given the per-part latent $\Phi_b(\mathbf{v}_b)$, the periodic parameters are extracted by applying a differentiable real-valued FFT along the time axis. Let $\{c_j\}$ denote the FFT coefficients of latent channel $k$. The amplitude, frequency, and offset are computed as
\begin{equation}
  A_{b,k} = \sqrt{\tfrac{2}{T}\sum_{j \ge 1}|c_j|^2}, \quad
  F_{b,k} = \frac{\sum_{j \ge 1} f_j |c_j|^2}{\sum_{j \ge 1}|c_j|^2}, \quad
  B_{b,k} = \tfrac{1}{T} c_0,
  \label{eq:fft-extraction}
\end{equation}
where $f_j$ is the FFT bin frequency. The phase shift $S_{b,k}$ is predicted by a separate fully-connected head followed by $\mathrm{arctan2}$, producing values in $(-\pi, \pi]$.

\textbf{Trunk Input with Root Linear Velocity.}
The root linear velocity $\mathbf{v}^{\text{root}}_{\text{lin}} \in \mathbb{R}^{T \times 3}$ exhibits clear periodicity during locomotion (vertical oscillation and forward-backward acceleration), and the phase branch should absorb it. We concatenate it to the trunk joint velocities: $\mathbf{v}_{\text{TK}}(\tau) = [\mathbf{v}^{\text{joint}}_{\text{TK}}(\tau);\, \mathbf{v}^{\text{root}}_{\text{lin}}(\tau)]$. The root yaw rate is omitted from the phase branch, since the pose branch already absorbs angular changes through the temporal evolution of $\mathbf{R}^{\text{joint}}$.

\textbf{Pelvis $v_x, v_z$ Soft Mask.}
Under the root-local frame, the pelvis horizontal velocity components $v_x, v_z$ are identically zero by definition. Treating them with a hard zero mask in $\mathcal{L}_{\text{phase}}$ caused training drift of approximately $\pm 0.8$ on neighboring channels. We instead apply a soft mask with weight $0.1$, which anchors these channels to zero while keeping a small but nonzero gradient signal. This stabilizes the trunk reconstruction without distorting the manifold.

\subsection{Cross-Embodiment Pipeline Details}
\label{app:cross_pipeline}

We provide details of the encoder and decoder adapters and the robot pose branch.

\textbf{Per-Robot Input Adapter $E_r^k$.}
The input adapter maps the robot velocity dimension $D_k$ to the human dimension 66 via a single global linear layer, $E_r^k = \text{Linear}(D_k \to 66)$. We considered a per-part variant with five separate linear layers, one per body part, but the two options produced near-identical retrieval and reconstruction quality. We adopt the global linear variant since it requires no hardcoded joint-to-part mapping and allows cross-joint information flow to be learned end-to-end.

\textbf{Forward Pipeline for Robots.}
Given a robot velocity input, the pipeline proceeds as follows. (1) Apply the per-robot input adapter: $\tilde{\mathbf{v}}^{(66)}_{R^k} = E_r^k(\mathbf{v}^{\text{joint}}_{R^k})$. (2) Split the 66-D output into the five-part dictionary using the same partition as the human. (3) Concatenate the robot's root linear velocity $\mathbf{v}^{\text{root}}_{\text{lin},k}$ to the trunk input. (4) Apply the frozen weight-shared encoder $E_X \equiv E_h$ to extract per-part periodic parameters. (5) Construct $\mathbf{P}_{R^k}(\tau) \in \mathbb{R}^{T \times 32}$ via \cref{eq:per_part_signal}.

\textbf{Decoder Split and Mirror Initialization.}
The human phase decoder $D_h$ remains frozen and is used only on the human path. Robot paths share a unified decoder $D_X$ that is initialized as a weight-copy of $D_h$ and fine-tuned on robot data. This split provides a gradient safety net: robot reconstruction loss flows only through $D_X$ and never disturbs $D_h$. The mirror initialization gives $D_X$ a strong starting point, since robot motion is itself a retargeted humanoid signal and the human velocity decoder serves as a valid prior. We also tested freezing $D_X$ at its mirror copy; fine-tuning improved retrieval mAP and reconstruction by approximately $1\%$ on average and was adopted as the default.

\textbf{Per-Robot Pose Branch.}
The pose branch is instantiated separately for each embodiment, since posture distributions and joint conventions vary across morphologies. \cref{tab:pose-branch-modules} summarizes which modules are shared with the human oracle and which are trainable per-embodiment. The phase-side modules ($\mathcal{E}_\psi$ and the phase-to-pose FiLM head) are shared and frozen, consistent with anchoring the manifold to the human reference. The pose-side modules ($\mathcal{E}_\phi$ and the pose-to-phase FiLM head) are per-embodiment trainable, since their input $\mathbf{z}_{\text{pose}}$ comes from an embodiment-specific encoder with a different distribution.

\begin{table}[h]
\centering
\small
\begin{tabular}{l c c}
\toprule
\textbf{Module} & \textbf{Human} & \textbf{Robot $k$} \\
\midrule
Pose encoder + decoder           & frozen               & per-embodiment, trainable \\
$\mathcal{E}_\phi$ (Pose$\to$Phase manifold enc.) & frozen   & per-embodiment, trainable \\
Pose$\to$Phase FiLM head         & frozen               & per-embodiment, trainable \\
$\mathcal{E}_\psi$ (Phase$\to$Pose manifold enc.) & shared, frozen & shared, frozen \\
Phase$\to$Pose FiLM head         & shared, frozen       & shared, frozen \\
\bottomrule
\end{tabular}
\vspace{4pt}
\caption{\textbf{Pose branch module sharing across embodiments.} Phase-side conditioning modules are shared with the frozen human oracle, while pose-side conditioning modules are trainable per-embodiment.}
\label{tab:pose-branch-modules}
\end{table}

\subsection{Alignment Objective Details}
\label{app:alignment}
\textbf{Alignment Head Architecture.}
The projection $\pi$ takes the per-frame concatenated feature $\mathbf{f}(\tau) = [\mathbf{P}(\tau);\, \mathbf{z}^{\text{pool}}_{\text{pose}}] \in \mathbb{R}^{160}$ and produces a unit-sphere embedding. Its structure is $\text{Linear}(160 \to 64) \to \text{GELU} \to \text{Linear}(64 \to 64) \to \ell_2$. Per-frame normalization yields the trajectory $z(\tau) \in \mathcal{S}^{63}$, and a subsequent time-pool with another $\ell_2$ normalization yields the clip-level descriptor $z^{\text{pool}} \in \mathcal{S}^{63}$. The trajectory $z(\tau)$ feeds $\mathcal{L}_{\text{speed}}$ and $\mathcal{L}_{\text{shape}}$, while $z^{\text{pool}}$ feeds $\mathcal{L}_{\text{pair}}$.

\textbf{Pose-Half Near-Zero Initialization.}
The first linear layer of $\pi$ has weight matrix $W \in \mathbb{R}^{64 \times 160}$. We initialize the first 32 columns (the phase input half) with standard Kaiming initialization and the remaining 128 columns (the pose input half) with a near-zero scale of $10^{-3}$. At the start of training, $\pi$ thus behaves close to a phase-only projection, and the pose contribution grows gradually through gradient updates. This warm-onset prevents pose noise from disrupting alignment early and keeps the pose contribution at an auxiliary level by the end of training.

\textbf{Hierarchical Pair Loss.}
We apply the LAMP soft target at two granularities. Let $\boldsymbol{\ell}^{\text{W}}_{\rightarrow}$ denote the $B \times B$ window-level logits over the human-to-robot or robot-to-human direction, and $T^{\text{soft}}_{\text{W}}$ the corresponding LAMP-induced soft target from per-window LAMP features. The window-level loss is
\begin{equation}
  \mathcal{L}^{(k)}_{\text{pair,W}} = \tfrac{1}{2} \sum_{\rightarrow} \mathrm{CE}\big(\boldsymbol{\ell}^{\text{W}}_{\rightarrow},\, (1 - \lambda_{\text{mix}}) T^{\text{hard}} + \lambda_{\text{mix}} T^{\text{soft}}_{\text{W}}\big).
  \label{eq:pair-W}
\end{equation}
For the clip-pool head, we average $z$ over windows sharing the same clip, normalize, and form a $K \times K$ logit matrix where $K$ is the number of unique clips in the batch. The corresponding LAMP soft target $T^{\text{soft}}_{\text{C}}$ uses clip-level LAMP features:
\begin{equation}
  \mathcal{L}^{(k)}_{\text{pair,C}} = \tfrac{1}{2} \sum_{\rightarrow} \mathrm{CE}\big(\boldsymbol{\ell}^{\text{C}}_{\rightarrow},\, T^{\text{soft}}_{\text{C}}\big).
  \label{eq:pair-C}
\end{equation}
The clip-pool head uses a pure soft target since same-clip windows are already merged into a single anchor through pooling, removing the gradient-dilution effect of broadcasting soft labels across redundant windows. The final pair loss combines both granularities:
\begin{equation}
  \mathcal{L}^{(k)}_{\text{pair}} = \mathcal{L}^{(k)}_{\text{pair,W}} + \lambda_{\text{C}} \mathcal{L}^{(k)}_{\text{pair,C}}.
  \label{eq:pair-hier}
\end{equation}

\textbf{Trajectory Consistency Terms.}
The two auxiliary geometric penalties on $z(\tau) \in \mathcal{S}^{63}$ are
\begin{equation}
  \mathcal{L}^{(k)}_{\text{speed}} = \tfrac{1}{T-1}\sum_{\tau=1}^{T-1} \big|\|z_H(\tau{+}1){-}z_H(\tau)\| - \|z_{R^k}(\tau{+}1){-}z_{R^k}(\tau)\|\big|,
  \label{eq:speed-loss}
\end{equation}
\begin{equation}
  \mathcal{L}^{(k)}_{\text{shape}} = \tfrac{2}{T}\sum_{\tau=1}^{T/2} \big|\|z_H(\tau{+}T/2){-}z_H(\tau)\| - \|z_{R^k}(\tau{+}T/2){-}z_{R^k}(\tau)\|\big|.
  \label{eq:shape-loss}
\end{equation}
$\mathcal{L}_{\text{speed}}$ matches per-frame step size on the sphere, an implicit frequency synchronization. $\mathcal{L}_{\text{shape}}$ matches half-period distance, an implicit amplitude synchronization. Both operate on the post-projection unit sphere, making them scale-invariant by construction.

\textbf{LAMP Teacher.}
LAMP~\citep{li2025lamp} is a pretrained motion-language prior trained on human motion. Its 512-D embedding serves as a fixed teacher. We use two granularities of LAMP features: clip-level embeddings for the clip-pool head, and stride-20 window-level embeddings for the window head. Both are precomputed on the human motion data and applied symmetrically as soft targets for human-to-robot and robot-to-human alignment in each batch.


\subsection{Training Procedure}
\label{app:training}

\paragraph{Reconstruction Loss Terms.}
The per-part velocity reconstruction loss applies a weighted MSE over the five body parts:
\begin{equation}
  \mathcal{L}^{(e)}_{\text{phase}} = \sum_{b \in \mathcal{B}} w_b \, \|\hat{\mathbf{v}}_b^{(e)} - \mathbf{v}_b^{(e)}\|_2^2,
\end{equation}
where the pelvis $v_x, v_z$ channels carry a soft weight $w = 0.1$ as described in \cref{app:phase_module}. The pose reconstruction loss combines a geodesic distance on 6D continuous rotations with a root position MSE:
\begin{equation}
  \mathcal{L}^{(e)}_{\text{pose}} = \mathrm{geodesic}(\hat{\mathbf{R}}^{\text{joint},(e)}, \mathbf{R}^{\text{joint},(e)}) + \|\hat{\mathbf{X}}^{\text{root},(e)} - \mathbf{X}^{\text{root},(e)}\|_2^2.
\end{equation}
The forward kinematics consistency loss links the two branches by enforcing agreement on global joint positions:
\begin{equation}
  \mathcal{L}^{(H)}_{\text{FK}} = \big\| \mathrm{FK}(\hat{\mathbf{R}}^{\text{joint}}, \hat{\mathbf{X}}^{\text{root}}) - \mathrm{FK}(\mathbf{R}^{\text{joint}}, \mathbf{X}^{\text{root}}) \big\|_2^2.
\end{equation}
The FK loss is applied only on the human path, since we do not maintain a differentiable kinematic chain for each robot in this work.

\textbf{Hyperparameters.}
\cref{tab:hyperparams} lists the full hyperparameter configuration. All experiments share these values across the human and four robot embodiments. 

\begin{table}[h]
\centering
\small
\begin{tabular}{l l c}
\toprule
\textbf{Group} & \textbf{Item} & \textbf{Value} \\
\midrule
Window & $T$, fps & 121, 60 \\
Phase  & $K_b$ (LA, RA, TK, LL, RL) & (3, 3, 2, 4, 4), $M_{\text{total}}=16$ \\
Pose   & $L_{\text{pose}}$, $d_{\text{pose}}$, $d_R$, $d_{\text{root}}$ & 8, 128, 96, 32 \\
FiLM   & $d_g$, $d_{g,\text{pose}}$, $\alpha$ init & 128, 64, 0.05 \\
Alignment & $\pi$ input dim, $d_{\text{align}}$ & 160 (= 32 + 128), 64 \\
       & pose-half init scale & $10^{-3}$ \\
       & $\tau_h$, $\tau_s$ & 0.1, 0.07 \\
       & $\lambda_{\text{mix}}$, $\lambda_{\text{C}}$ & 0.3, 0.3 \\
       & $\lambda_{\text{pair}}$, $\lambda_{\text{speed}}$, $\lambda_{\text{shape}}$ & 1.0, 0.5, 0.5 \\
Loss weights & $\lambda_{\text{phase}}$, $\lambda_{\text{pose}}$, $\lambda_{\text{FK}}$ & 5.0, 5.0, 1.0 \\
             & pelvis $v_x, v_z$ soft mask & 0.1 \\
Optimization & optimizer, lr, weight decay & AdamW, $2 \times 10^{-4}$, $10^{-4}$ \\
             & batch size, grad clip & 512, 5.0 \\
             & epochs (each stage) & 30 \\
             & FK warmup, alignment warmup & 3, 5 epochs \\
\bottomrule
\end{tabular}
\vspace{4pt}
\caption{\textbf{Hyperparameter configuration} for both training stages.}
\label{tab:hyperparams}
\end{table}

\textbf{Parameter Accounting.}
At Stage 2, the human oracle contributes approximately 528K frozen parameters, including the phase encoder body $E_h$ (shared as $E_X$ on the robot path), the manifold encoder $\mathcal{E}_\psi$, the phase-to-pose FiLM head, the phase decoder $D_h$, the human pose encoder and decoder, and the human pose-to-phase modules. The trainable budget is approximately 1.46M parameters, consisting of the shared phase decoder $D_X$ (~51K, mirror-initialized from $D_h$), the alignment head $\pi$ (~10K), per-robot input and output adapters $E_r^k, D_r^k$ (~11K each), per-robot pose encoder and decoder (~225K each), per-robot $\mathcal{E}_\phi$ (~43K each), and per-robot pose-to-phase FiLM head (~87K each), summed over the four robot embodiments.


\section{Additional Experiments}

\subsection{Ablation}
\label{sec:exp_ablation}

We isolate each design choice through clip-level retrieval on Stage~2 (\cref{tab:retrieval_clip}). Clip-level matching only requires preserving coarse \emph{semantic} identity, not fine temporal structure, making it an easy task: all methods exceed $91\%$ mAP and gaps between variants are small. We read this as a sanity check that each variant retains clip-level semantics, while our true advantage appears in exact-match retrieval (\cref{sec:exp_retrieval}), addressing the limitation in \cref{sec:conclusion}.

\begin{table*}[t]
\centering
\label{tab:retrieval_clip}
\setlength{\tabcolsep}{3pt}
\renewcommand{\arraystretch}{1.05}
\small
\begin{tabular}{l | cccc | cccc | cccc}
\toprule
 & \multicolumn{4}{c|}{\textbf{H$\to$R Clip-level}} & \multicolumn{4}{c|}{\textbf{R$\to$H Clip-level}} & \multicolumn{4}{c}{\textbf{R$\to$R Clip-level}} \\
\cmidrule(lr){2-5}\cmidrule(lr){6-9}\cmidrule(lr){10-13}
Method & R@1 & R@5 & R@10 & mAP & R@1 & R@5 & R@10 & mAP & R@1 & R@5 & R@10 & mAP \\
\midrule
MLP   & \textbf{99.6} & \textbf{100.0} & \textbf{100.0} & \textbf{99.8} & \textbf{99.8} & \textbf{100.0} & \textbf{100.0} & \textbf{99.9} & \textbf{99.0} & \textbf{99.8} & \textbf{100.0} & \textbf{99.4} \\
SONIC & 97.8 & 99.8 & 99.8 & 98.7 & 97.0 & 99.8 & 99.8 & 98.3 & 97.2 & \underline{99.5} & 99.5 & 98.2 \\
\midrule
Only Phase  & 93.8 & 98.1 & 98.8 & 95.7 & 94.2 & 98.5 & 99.0 & 96.2 & 88.1 & 96.0 & 98.0 & 91.4 \\
W.O LAMP    & 95.8 & 99.1 & 99.8 & 97.2 & 95.5 & 99.0 & 99.8 & 97.0 & 91.4 & 97.8 & 99.0 & 94.2 \\
W Pos       & 97.5 & \underline{99.9} & \textbf{100.0} & 98.5 & 98.8 & 99.8 & \textbf{100.0} & 99.3 & 93.5 & 99.3 & \underline{99.7} & 96.0 \\
\midrule
\textbf{W $Z_{\text{pos}}$ (Ours)} & \underline{99.1} & 99.8 & \underline{99.8} & \underline{99.4} & \underline{99.5} & \underline{99.8} & \underline{99.8} & \underline{99.6} & \underline{98.1} & 99.4 & \underline{99.7} & \underline{98.6} \\
\bottomrule
\end{tabular}
\caption{Cross-embodiment clip-level retrieval on the DPAE val set (2{,}472 clips).
  \textbf{H$\to$R}: human query $\to$ robot gallery (avg.\ over G1/H1.2/T1/BHL).
  \textbf{R$\to$H}: robot query $\to$ human gallery (avg.\ over G1/H1.2/T1/BHL).
  \textbf{R$\to$R}: robot $\leftrightarrow$ robot (avg.\ over $\binom{4}{2}{=}6$ pairs, both directions).
  All values in \%.
  \textbf{Bold}: best per column. \underline{Underline}: second best.}
\end{table*}

\textbf{Only Phase} (phase manifold alone) already reaches $91$--$96\%$ mAP, confirming the periodic prior captures clip identity on its own; adding LaMP guidance (\textbf{W.O LAMP}$\to$default) and root position (\textbf{W Pos}) each gives a small, consistent gain, largest on R$\to$R. Our full model \textbf{W $Z_{\text{pos}}$} is essentially tied with the best configuration everywhere. The MLP and VQ baselines lead at clip level, as expected: an unstructured objective directly optimizes coarse semantic identity, whereas our embedding additionally encodes \emph{when} each chunk occurs, which is unnecessary here but drives our large margin in exact-match. Adding temporal structure thus costs almost nothing on this easy task while dominating the fine-grained one.

\subsection{Effect of the LAMP Mixing Ratio}
\label{app:lamp_ratio}

\begin{figure}[h]
\centering
\includegraphics[width=\linewidth]{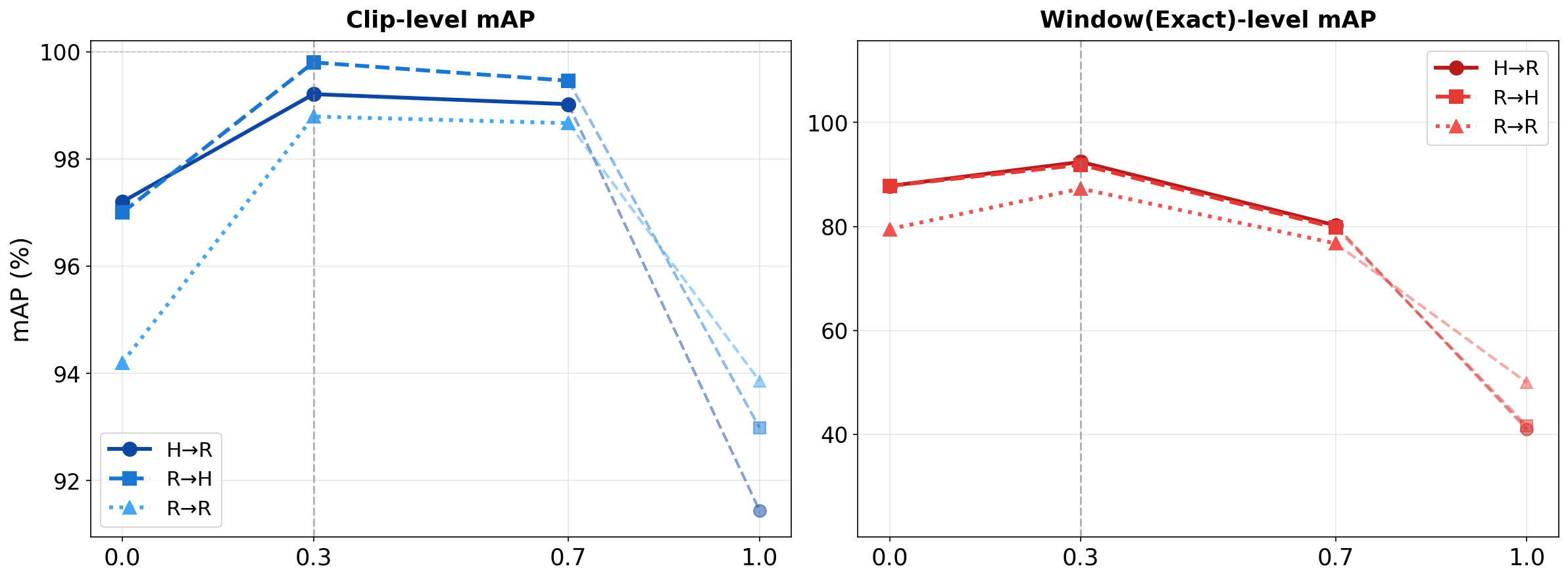}
\caption{\textbf{Effect of the LAMP mixing ratio $\lambda_{\text{mix}}$.} Clip-level (left) and exact-match (right) mAP over $\lambda_{\text{mix}} \in \{0.0, 0.3, 0.7, 1.0\}$ for all three directions. Both peak at $0.3$ (our default) and collapse at $1.0$ when the hard target is removed.}
\label{fig:lamp_ratio}
\end{figure}

The hierarchical pair loss (\cref{eq:pair-W}) mixes a hard-target InfoNCE term with a LAMP-induced soft target via $\lambda_{\text{mix}}$: at $\lambda_{\text{mix}}{=}0$ it is pure InfoNCE, at $\lambda_{\text{mix}}{=}1$ it relies entirely on LAMP semantic similarity. We sweep $\lambda_{\text{mix}} \in \{0.0, 0.3, 0.7, 1.0\}$ and report clip-level and exact-match mAP across the three directions (\cref{fig:lamp_ratio}).

Both granularities peak at $\lambda_{\text{mix}}{=}0.3$, our default. A moderate semantic signal helps, most visibly on the unsupervised R$\to$R direction where it strengthens transitive alignment through the human anchor. But beyond $0.3$ performance degrades, collapsing at $\lambda_{\text{mix}}{=}1.0$: with the hard target removed, the model loses the instance-level correspondence needed for fine-grained matching, since LAMP similarity alone is too coarse to separate chunks within a clip. LAMP thus works best as auxiliary guidance, not a standalone target.

\subsection{Downstream Tasks}
\label{sec:exp_downstream}

We probe the frozen embedding's utility in three settings: single-embodiment next-frame prediction (imitation), cross-embodiment teleoperation, and locomotion RL with a phase-shaped reward. In both offline tasks the policy head is a 3-layer MLP ($256\to256\to$output) trained for 50 epochs (Adam, lr $10^{-4}$, cosine schedule, batch 512). A \textbf{no-embedding baseline~(A)} receiving only raw proprioceptive state is included for reference.

\subsection{RL Locomotion Details}
\label{app:rl_reward}

\subsubsection{Reference Embedding Preparation}

Before RL training, every AMASS reference clip is encoded offline into per-frame reference embeddings $z_\text{ref}$ by each model's frozen human encoder, yielding a 64D sequence for PHASOR, 10D for VQ, and 80D for MLP. Each variant's reward uses its own pre-encoded $z_\text{ref}$, so the human encoder need not run during training.

\subsubsection{Environment Configuration}

\begin{itemize}[nosep]
\item Simulator: IsaacLab with GPU physics.
\item Simulation timestep $1/120$\,s, control decimation 2, giving a policy timestep of $1/60$\,s.
\item 8{,}192 parallel environments.
\item Episode length: up to 20\,s ($1{,}200$ policy steps).
\item Each episode resets to a standing pose (no reference-state initialization).
\end{itemize}

\subsubsection{Observation}

The policy observation is a 10-step history of proprioceptive state, concatenating per step the joint positions (23), joint velocities (23), base angular velocity (3), base roll/pitch/yaw (3), and previous action (23), for a total of 750 dimensions. This is augmented with the frozen robot embedding $z_{\text{align\_r}}$, whose dimension depends on the encoder:

\begin{center}
\small
\begin{tabular}{lcc}
\toprule
Variant & $z_{\text{align\_r}}$ dim & Total obs dim \\
\midrule
PHASOR & 64 & 814 \\
MLP    & 80 & 830 \\
VQ     & 10 & 760 \\
\bottomrule
\end{tabular}
\end{center}

The robot embedding is computed online: at each step the latest joint velocities are appended to a 121-frame sliding window, and once the window is full the frozen robot encoder produces $z_{\text{align\_r}}$.

\subsubsection{Reference Progression}

At each reset a reference clip is sampled at random, and its frame index advances by one per policy step (wrapping at the clip end). The corresponding reference embedding $z_{\text{ref\_cur}}$ is read from the precomputed sequence. The reference is used \emph{only} to compute the reward and is never part of the observation.

\subsubsection{Reward Design (15 Terms)}

\textbf{Survival (always active):}
\begin{center}
\small
\setlength{\tabcolsep}{4pt}
\begin{tabular}{clc}
\toprule
\# & Term & Weight \\
\midrule
1  & Termination penalty (torso height $< 0.3$\,m) & $-200$ \\
2  & Alive bonus ($+0.15$ per step) & $\times 1$ \\
3  & Flat orientation: $-(g_{b,x}^2 + g_{b,y}^2)$ & $-1.0$ \\
4  & Base height: $-(z_\text{pelvis} - 0.78)^2$ & $-10.0$ \\
5  & Vertical velocity: $-v_{z}^2$ & $-2.0$ \\
6  & Roll/pitch angular velocity & $-0.05$ \\
\bottomrule
\end{tabular}
\end{center}

\textbf{Locomotion (active from iteration 500):}
\begin{center}
\small
\setlength{\tabcolsep}{4pt}
\begin{tabular}{clc}
\toprule
\# & Term & Weight \\
\midrule
7  & Velocity tracking: $\exp(-(v_\text{robot} - v_\text{ref})^2 / 0.25)$ & $\lambda_\text{vel}$ \\
8  & Feet air-time (bipedal alternation bonus) & $+2.0$ \\
9  & Feet slide penalty & $-0.1$ \\
10 & Hip joint deviation from default & $-0.5$ \\
11 & Arm joint deviation from default & $-0.1$ \\
12 & Torso joint deviation from default & $-0.1$ \\
13 & Ankle soft-limit violation & $-1.0$ \\
14 & Joint acceleration (hip, knee) & $-1.25{\times}10^{-7}$ \\
15 & Joint torque (hip, knee, ankle) & $-1.5{\times}10^{-7}$ \\
16 & Action rate: $-(a_t - a_{t-1})^2$ & $-0.001$ \\
\bottomrule
\end{tabular}
\end{center}

\textbf{Phase (active from iteration 1000):}
\begin{center}
\small
\begin{tabular}{clc}
\toprule
\# & Term & Weight \\
\midrule
17 & Phase similarity: $(\cos(z_{\text{align\_r}}, z_{\text{ref\_cur}}) + 1)/2$ & $\lambda_\text{phase}$ \\
\bottomrule
\end{tabular}
\end{center}

Non-termination terms are clamped to be non-negative before the termination penalty is added, so the agent cannot reduce its loss by ending an episode early; the termination penalty therefore always applies.

\subsubsection{PPO Configuration}

\begin{itemize}[nosep]

\item 5 PPO epochs, 4 mini-batches per iteration; 2{,}500 iterations total.
\item Policy network: MLP $[512 \to 256 \to 128]$ with ELU, observation normalization, and Gaussian action noise $\sigma = 0.3$.
\item Curriculum weights updated every 50 iterations.
\end{itemize}

\subsubsection{Curriculum Schedule}

\begin{center}
\small
\begin{tabular}{lccl}
\toprule
Iteration & $\lambda_\text{vel}$ & $\lambda_\text{phase}$ & Stage \\
\midrule
0--499     & 0 & 0         & Survival only (stand) \\
500--999   & 1 & 0         & $+$ velocity (walk) \\
1000--1999 & 1 & $0 \to 1$ & $+$ phase ramp (shape gait) \\
2000--2500 & 1 & 1         & Full reward \\
\bottomrule
\end{tabular}
\end{center}

\end{document}